# Interactive Attention AI to translate low light photos to captions for night scene understanding in women safety

A. Rajagopal[1], V. Nirmala[2], Arun Muthuraj Vedamanickam[3]

[1] Indian Institute of Technology, Madras, India
[2] Queen Marys College, Chennai, India (corresponance)
[3] National Institute of Technology, India
[1]`rajagopal.motivate@gmail.com`, [2]`gvan.nirmala@gmail.com`,
[3]`arun.gvan@gmail.com`

**Abstract.** There is amazing progress in Deep Learning based models for Image captioning and Low Light image enhancement. For the first time in literature, this paper develops a Deep Learning model that translates night scenes to sentences, opening new possibilities for AI applications in the safety of visually impaired women. Inspired by Image Captioning and Visual Question Answering, a novel "Interactive Image Captioning" is developed. A user can make the AI focus on any chosen person of interest by influencing the attention scoring. Attention context vectors are computed from CNN feature vectors and user-provided start word. The Encoder-Attention-Decoder neural network learns to produce captions from low brightness images. This paper demonstrates how women safety can be enabled by researching a novel AI capability in the Interactive Vision-Language model for perception of the environment in the night.

## 1 Introduction

### 1.1 Need & Significance

The need to research scene understanding at night is important for developing Deep Learning applications for assisting the visually impaired and women safety. In a scenario of a visual impaired girl straying into a vulnerable street at night, a real-time audio description of the scene from her smartphone camera can help her to perceive the scene. Further in the scenario, she can ask the AI to focus its Attention on an a salient object, motivating the need for a novel idea of *"AI based Interactive Captioning of night scenes with user guided attention mechanism"*.

Starting with the need to develop girl safety apps, arises the need for research on the development of Vision-Language models that can perceive night scenes for on-device ML inference on smartphones or wearables. The AI has to be to functional on visual impaired user's smartphone even when there is no internet. Also the on-device ML inference has to be fast enough for real-time interaction with the user. This puts a boundary on the size of the neural network, hence the need for research. The multi-modal embedding approach of multi-layer Transformer model OSCAR is a powerful idea in Vision-Language modelling [1]. It helps achieve state-of-art performance, but the challenge is in on-device ML inference for such gigantic models. Hence, there is a need to explore.

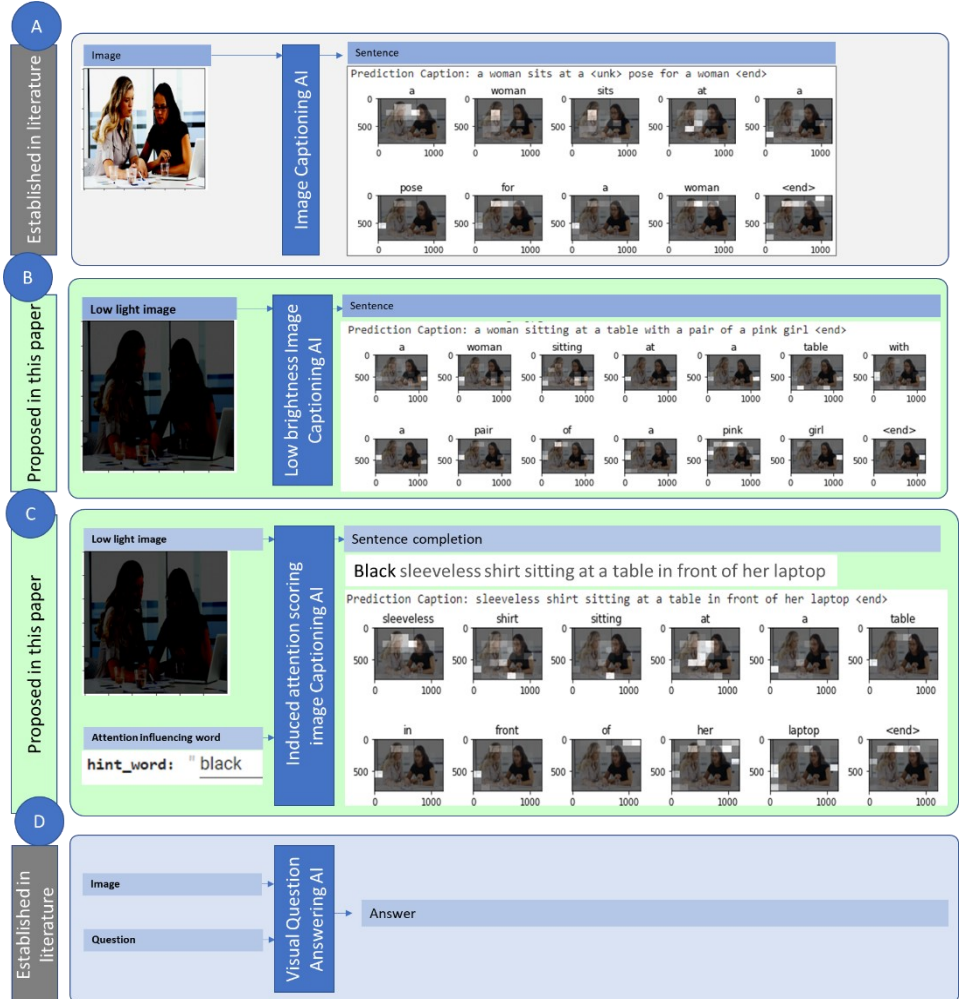

*Figure 1: Key contributions and novelty*

## 1.2 Contributions

Three areas contributed in this paper are
1) To the best of our knowledge, this is the 1st work to demonstrate the ability of a neural network to perceive night scenes and translate the low brightness image to text.

2) Propose "Interactive Captioning", where the user can induce the AI to focus its Attention on user-specified object.

3) Apply the proposed neural network for the safety of women and blind people.

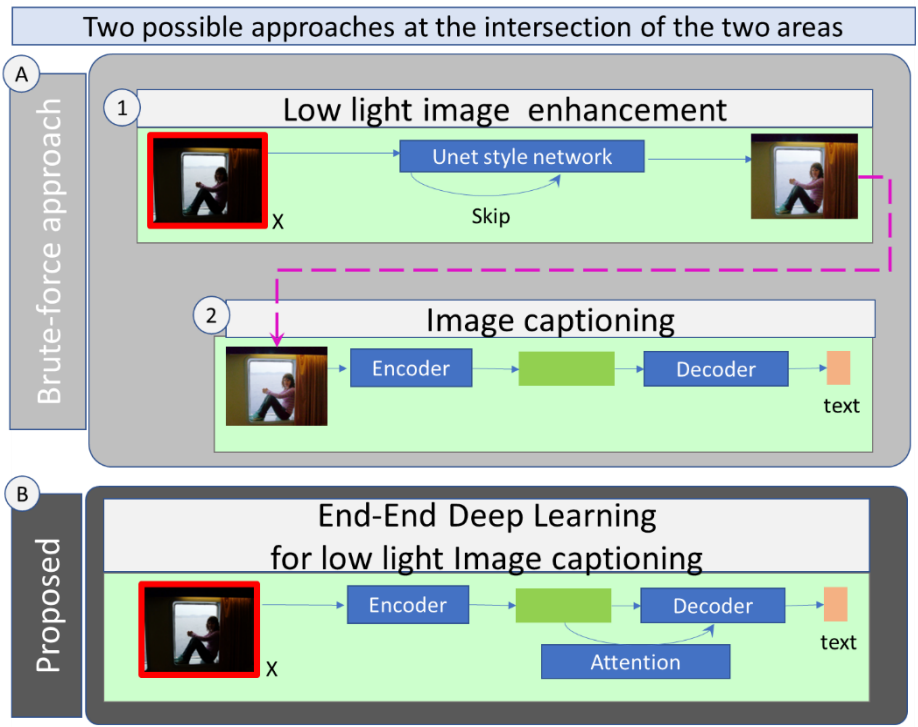

*Figure 2: Research gap & Architecture choices in fulfill the gap*

For the first time in literature, this paper proposes and implements a novel scene understanding AI that has the following unique contributions

1) AI's ability to see and understand photos taken in low light or night has applications in girl safety use cases. Hence, the first contribution is developing a deep learning-based visual language model to understand night scenes or photos taken in low light settings. Specifically, this paper develops and demonstrates that a Deep Learning model can translate any low brightness image into a sentence. The proposed model architecture is shown in contrast with a brute-force approach in Fig 3. A brute-force approach is a pipeline of first enhancing the image, and then making inferences with an image captioning network. In this paper, we use end-to-end Deep Learning to train an Encoder-Attention-Decoder on low-brightness images. Once the trained model is developed, we experiment with randomly downloaded images from the web. For example, a random nighttime image from the internet is translated into text as shown in Fig 2. These experiments demonstrate the ability of this neural network model to describe any night scene, inspite of low brightness in the image.

2) Attention modelling [7] is a powerful paradigm in Deep Learning and has been the foundation for much of the recent progress in the field. Transformers [1, 17] based approaches are entirely based on Attention modelling. In the "Attention is all you need" paper, Google Brain argued why modelling of Attention can train the neural networks similar to how humans focus their Attention of a particular salient object. While the usual norm is to allow the Image captioning neural network to generate the entire sentence, this paper develops a new variation, as shown in Fig 1.C. In this variation, users can provide a start word and inductively change the attention scores. Thus, experiments show the Image Captioning network can focus its Attention on a particular object chosen by the user and base the sentence on the salient object chosen by the user. This is the first time in literature, Interactive Caption on night scene comprehension is demonstrated. This new concept is shown in Fig 3.

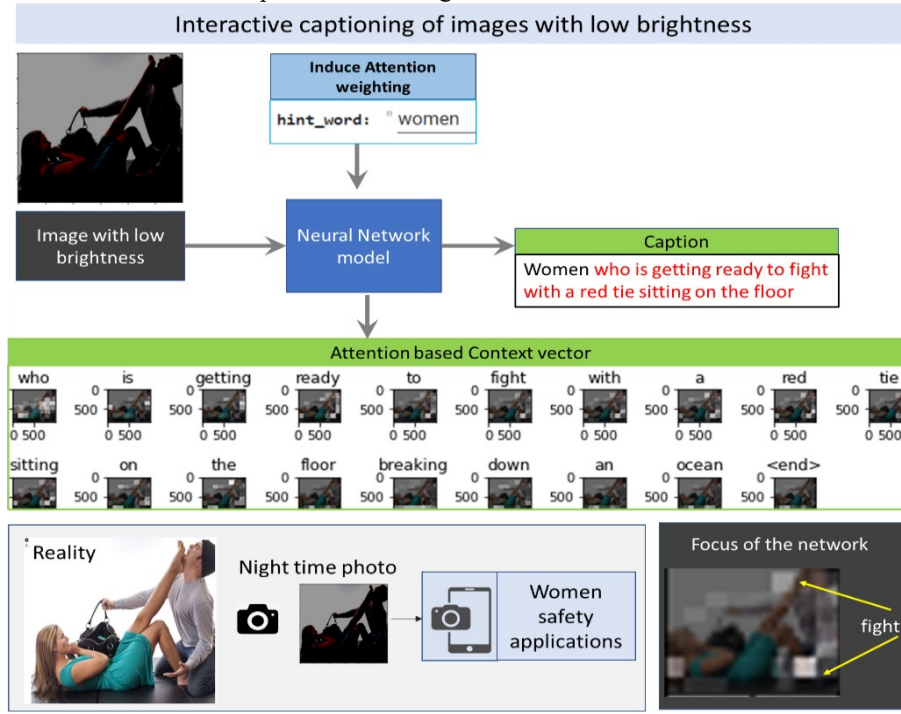

*Figure 3: Women safety applications of the proposed concept*

## 1.3 Novelty

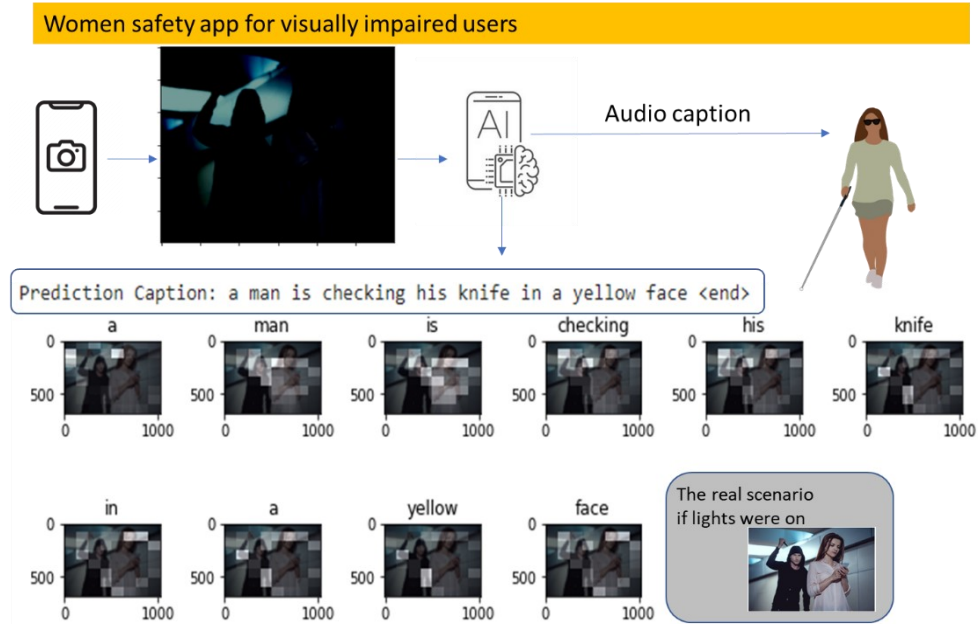

*Figure 4: People with vision impairments can use the proposed AI*

The research gap is articulated in Fig 2. There is amazing progress in two tasks
- Image Captioning & Visual Question Answering
- Low light image enhancement & super-resolution

Inspired by these advances, this paper develops a novel idea & demonstrates it. The novelty aspects is illustrated in Fig 1.
  A. Image captioning        (established)
  B. Low light Image captioning  *(proposed)*
  C. Interactive captioning     *(proposed)*
  D. Visual Question Answering  (established)

While Image Captioning and Visual Question Answering are well established in the literature, this paper contributes two novel areas. One is Image captioning of photos when the illumination is low. Another is implementing the variant of the concept of Visual Question Answering (VQA).

There are three aspects of novelty in this work
1. Novel Deep Learning modelling (Fig 1)
2. The idea of Interactive Captioning (Fig 6 & Fig 7)
3. A novel application for girl safety (Fig 4 & Fig 6)

### 1.4 Research gap & Related work

Since this paper is at the intersection of Low light image enhancement and Vision & Language interaction modelling [1], this section presents research in both these areas. While there is significant progress in both these areas, there are fewer publications at the intersection of these areas. This paper is at the intersection of these two areas. The literature in both these Image Caption & Image Enhancement tasks can be generalised and represented by a generic architecture pattern, as illustrated in Fig 2.

The literature study shows that while there is significant results in both these tasks of image enhancement & captioning, the gap in research is to combine the two tasks into a single network for efficient on-device ML. Two different approaches for modelling of Image captioning of night scenes is illustrated in Fig 2. The two approaches are
1. Brute-force approach: Combining the advancements in both Image Enhancement & Image captioning can be achieved by a simple pipeline architecture such as the one shown in Fig 2A. However, the real-time computation for on-device ML could be high with such a pipleline based approach.
2. Proposed approach: As per the trends of end to end Deep Learning based training, this paper takes an approach as shown in Fig 2B. Here images with low brightness is directly used as an input to the neural network

Deep Learning based Image enhancement is used in de-noising auto encoders, super-resolution, low light image enhancement. At the heart of image enhancement [8], an UNet encoder-decoder style architecture pattern is utilized to transform the image from one representation to another, as shown in Fig 2. There have been significant advancements in Image Super Resolution and Video Super resolution, certainly indicating the potential of CNNs for learning patterns and transforming them. Skip connections along with CNN layers transform different features at various levels. The U-Net inspired architecture is typically well established for image-image translation [10].

Deep Learning in combining Vision and Language was seen tremendous progress, especially with the latest achievement in human level parity in image captioning tasks. The multi-layer Transformer architecture OSCAR can transform multi-modal embedding has shown promising a new direction [1] However, the challenge for on-device ML requires a much smaller network given resource-constrained wearables and smartphone devices. Hence the need for research on compact versions of Vision & Language modelling. The success in image captioning arises from 3 factors
1. Multimodal embedding rather than just concatenation visual and word embeddings
2. Attention context vectors that compute the Attention to various salient objects in the image
3. Identifying the pairing of classes found in a image, and words in the caption

The various architecture patterns in Vision & Language modelling literature is summarized in Fig 5

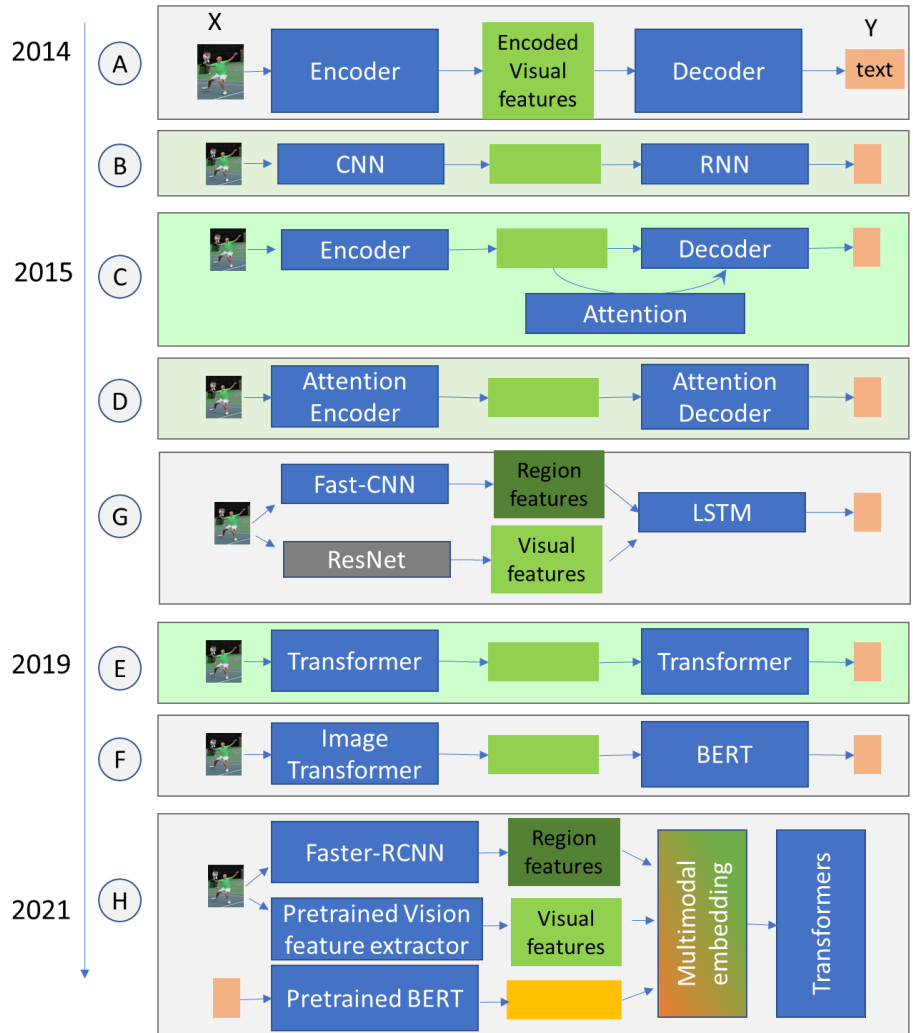

*Figure 5: Architecture generalized from Literature*

To generalize, most of the Vision-Language translation tasks such as Image Captioning can be performed by these approaches

1. Encoder-decoder architecture pattern (2014):
• Here the visual features are encoded by CNN & then the thought vector is then decoded by RNN. The classic paper is "Show & Tell" by Bengio[5]. The pattern is depicted in Fig 5 A. Refer Fig 5 B for this neural image captioning.

2. Encoder-Attention-Decoder (Since 2015):
• Since the advent of sequence to sequence neural machine translation, a popular choice of architecture is to apply different forms of attention technique. Significant progress was possible due to Attention in such translation tasks. This architecture pattern is shown in Fig 5C. "Show, Tell & Attend" paper [2] demonstrated Attention based image captioning.

3. Visual feature extractor in the architecture (2015-2021):

• The use of a IMAGNET based pre-trained feature extractor based in Image Captioning networks has found significant adoption till date. The most recent one have been with ResNet, though Inception is used in many cases. The pretrained Inception first extract visual features and then inputs to the encoder. [1]

4. Region feature extractor in the architecture (Since 2016):

• The use of object detectors with R-CNN framework has added localization data into addition to visual features. This is also used in OSCAR as shown in Fig 5G and Fig 5H.

5. Transformer based approach (2019-2021): With the advent of multi-head Attention, the use of Transformers for Encoding and Decoding has been popular in language modelling. For example, OSCAR uses transformers both for modelling and for text embedding using pre-trained BERT language models. (Ref Fig 5E and Fig 5H)

6. Multi-modal embedding & Transformer (2020-2021):

• OSCAR[1] relies its success based on the idea of multi-modal embedding and multi-layered multi-head Attention. (Ref Fig 5H)

## 2    AI architecture, Methods & Results

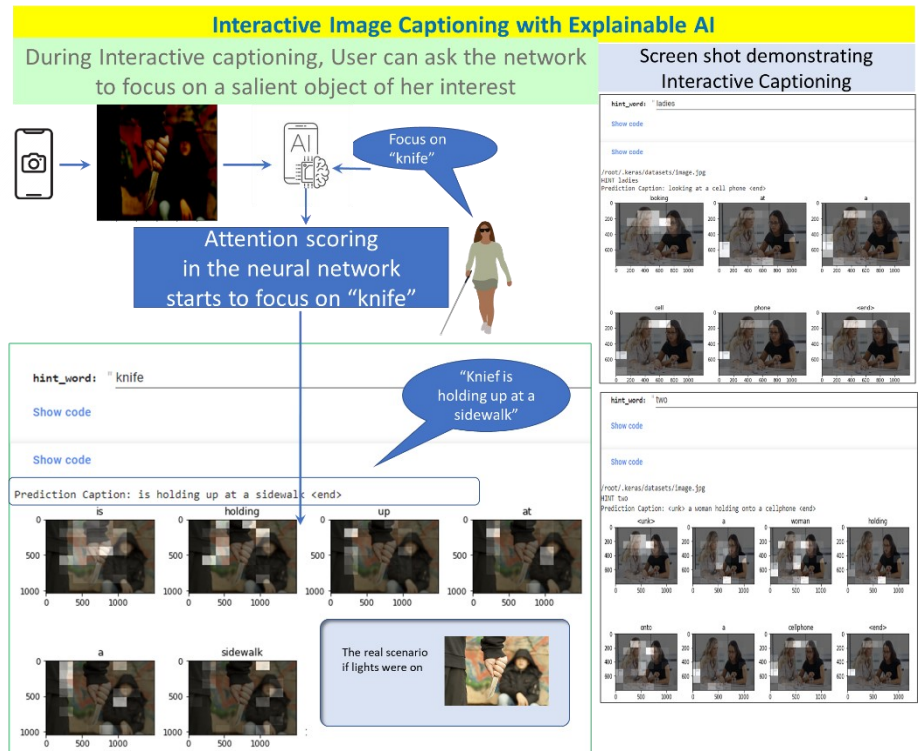

*Figure 6: Literature's 1st work on Interactive Image Captioning for night scene understanding*

## 2.1 AI for describing night scenes for visually impaired users

This paper assumes significance because it opens doors to life-saving applications for women & the visual impaired community. If one stands in the shoe of a visually impaired woman, the importance of developing this capability can be soundly defended. The significance of this work is in enabling safety use cases for visually impaired users (as illustrated in Fig 4 & Fig 6).

## 2.2 Novel result: This Neural Network translates low brightness images to sentences!

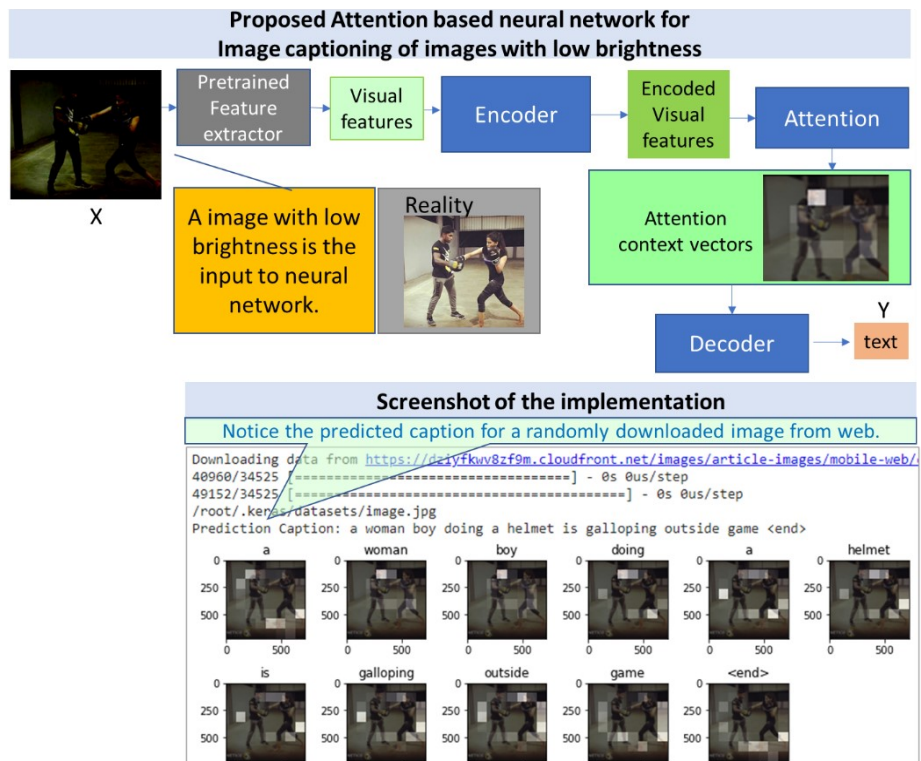

*Figure 7: Neural Network architecture*

The illumination level in the dark combined with exposure levels on camera in consumer devices means the quality of photos may not be comprehend-able clearly by human eyes. Yet, can machines comprehend such photos taken in low light. The proposed AI in this paper is shown to translate night scenes to sentences that describe the scene. The experimental results show this trained model can easily provide a caption for any randomly downloaded image from the internet. A screenshot of this capability is shown in Fig 7. Also, this capability to translate a random night scene into a sentence is demonstrated online.

The proposed neural network architecture is inspired by Encode-Attention-Decode architecture established in "Shown, Attend and Tell" paper. However, this feat of captioning night scenes is made possible by two novel ideas introduced in this paper. First, the Deep Learning model is trained on a modified MS-COCO dataset. Second, the idea of interactive user input directly influences Attention score computations. These two ideas are elaborated on in later sections.

## 2.3 Dataset

To address the non-availability of dataset to train image captioning for night environments, the paper synthetically creates a dataset from MS-COCO dataset, but adjusting the brightness of all the images. Thus the new dataset is derived from MS-COCO dataset [5]. The dataset now consists of modified images from MS-COCO dataset along with the corresponding captions provided in the MS-COCO dataset. The ddataset used for training comprises of {images with low brightness, 5 textual captions}. The modified dataset used in training the neural network is described in Fig 8.

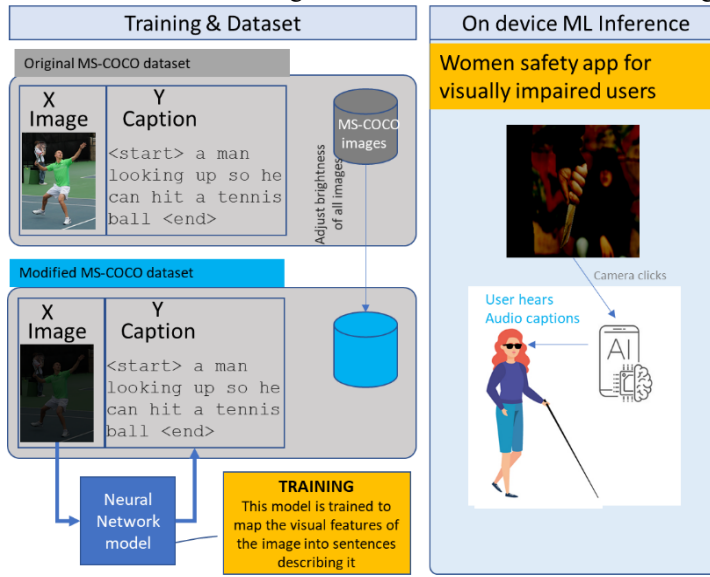

*Figure 8: Dataset preparation*

## 2.4 Novelty in AI inference for Interactive Caption Generation

While the classical concept of Deep Learning based automated caption generation is shown in Fig 1 A, the proposed concept of Interactive Caption Generation is shown in Fig 1C. A detailed illustration of the user experience by visual impaired women is in Fig 6. In this scenario, where you wish to "instruct" the AI to focus its Attention on particular object of interest, the user can simply ask the AI to include the user specified object in the caption to be generated. This is in contrast with the VQA model (Fig 1D). The VQA & the proposed concept require different dataset structures (refer Table 1).

|  | Training Dataset: | AI app Inference I/O |
|---|---|---|
| VQA | • X: Image, Question<br>• Y: Answer | • Inputs: Image, Question<br>• Output: Predicted Answer |
| Inter-active caption | • X: Low brightness Image<br>• Y: Sentence | • Inputs: Low light photo, word<br>• Output: Sentence completion |

*Table 1*

## 2.5   HMI in Attention model for Interactive Caption Generation

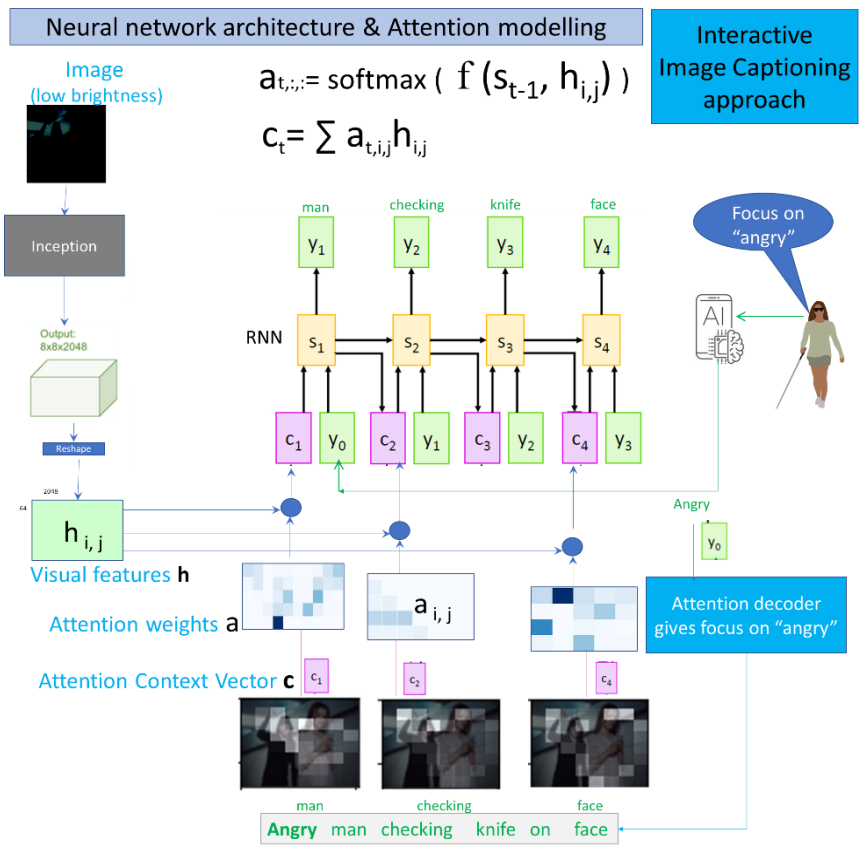

*Figure 9: Human in loop in Attention Decoder for Interactive captioning of night images*

In addition to proposing & developing the Deep Learning model for translation of night imagery to textual sentences, the paper explored the novel idea at the intersection of Human-Machine Interaction (HMI) and on-device ML inference for Image Captioning use case. The architecture for injecting user input to the Attention computation is presented in Fig 9. As seen in this architecture, when the user input is combined with the visual features, the attention context vector focuses on the salient object provided in the user input. Thus the Attention scoring is influenced by the human input, allowing for human in the loop intervention in the Attention scoring based caption generation mechanism. To result of this human input can be seen in Fig 6, where the screenshot shows the output generated depends upon the input word provided by the user.

The architecture consists of a visual feature extraction using the entire convolution layers of the IMAGENET pre-trained Inception model, and then computation of Attention alignments based on a combination of user input word and the visual features. By using attention computation such as dot product Attention or Bahdanau Attention, the Attention context vector (c) is computed from the learnt attention weights (a) and the visual embeddings (h). The Attention RNN then focus its Attention to generate the text description. The novelty is on this architecture in injecting the user typed word into this Attention computation architecture. The experiments shows promising results as shown in screenshots in Fig 6 and the online demo at URL. https://sites.google.com/view/low-lightimagecaption

## 2.6  Feasibility towards enabling women safety apps

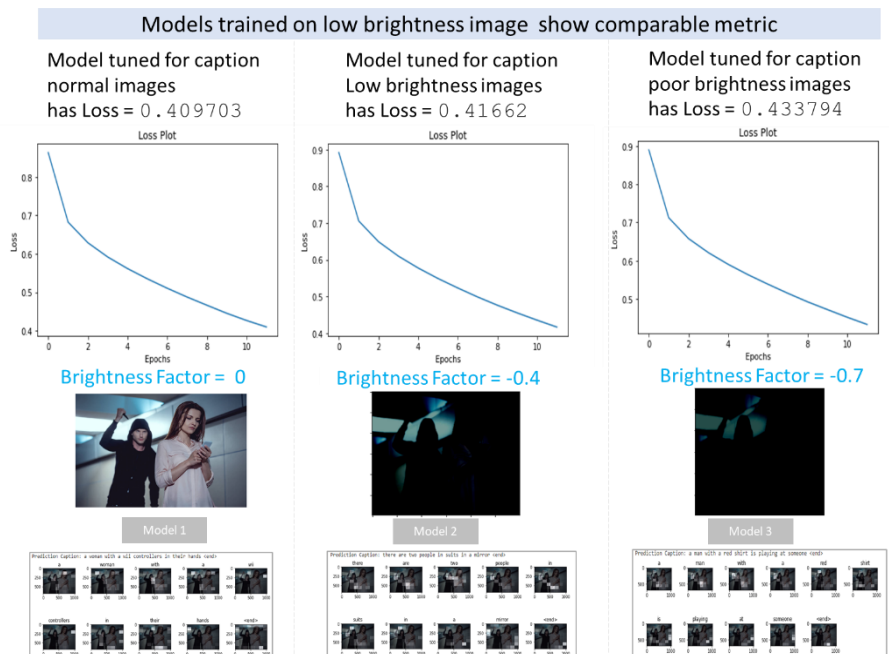

*Figure 10: Comparison of 3 different models trained of different environments*

The feasibility of developing women safety apps is demonstrated in this paper. The potential of employing a Deep Learning based approach to build Visual Language models that is capable of comprehending night scenes is established experimentally in this work. Early results show that the night scene captioning network network offers a similar level of accuracy/loss as compared to captioning network trained on MS-COCO. The experimental results on loss at 3 different networks are shown in Fig 10. The experiments shows loss is within a reasonable range (less than 5%) when compared to network trained on MSCOCO. Though this is a first baby step in this direction of safety applications for visual impaired women, a lot more research is required in the future to strengthen this idea. There is a plethora of opportunities for the research community to research more advanced AI ideas for enabling women safety applications. This paper makes the source code in open source at URL <URL is mased for double blind review>.

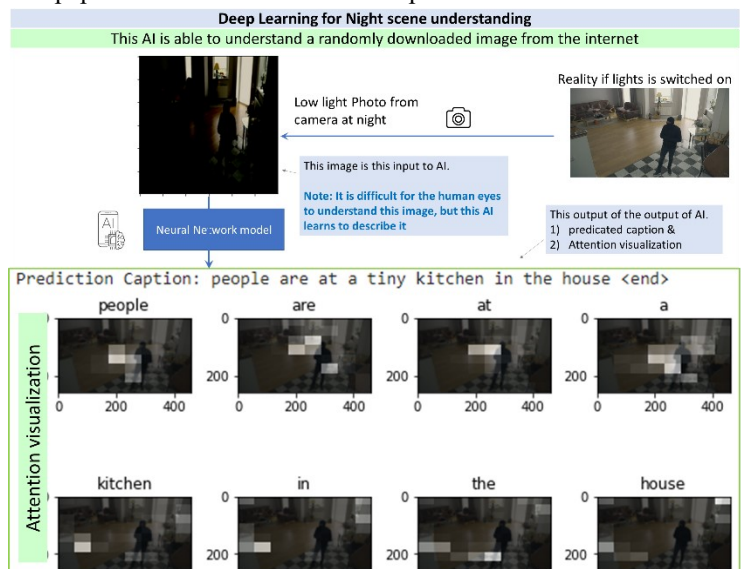

*Figure 11: Attention visualization*

## 3 Conclusions

The paper identified a unique challenge in developing Vision-Language Modelling for the understanding of low light night scenes. We motivated the need for the development of Deep Learning based Interactive Image Captioning models that can understand low light scenes. Smartphone AI based night scene perception is a crucial capability for personal safety apps for visually impaired women & girls.

Based on a review of literature, we identified the opportunity to contribute to the literature at the intersection of two areas, namely Image Enhancement & Image Captioning. Though each of these two topics has witnessed amazing progress, there are not many publications at the intersection of both Image Captioning & Low Light Image Enhancement. Based on the literature review, the generic pattern in architecture for Image Captioning is identified as Encoder-Decoder with Attention mechanism at the heart of most papers, with the recent focus on multi-head Attention & multimodal embeddings to combine Visual & Language modalities. Similarly, UNet style Encoder-Decoder architecture is found at the heart of Image Enhancement and Super-Resolution architectures. The role of Attention based weighting of visual features in an Encoder-Decoder architecture, and multimodal embeddings have shown promise in Vision & Language modelling tasks such as Image Captioning. Since girl safety applications require on-device

ML on resource constrained smartphones for real-time AI, the choice of neural network architecture requires careful investigation. Though Transformers based multimodal embedding architecture offer state-of-art captioning, its distillation is needed for affordance on smartphones due for the computation and memory requirements of multi-layer Transformers inspired models such as OSCAR. Hence, the need to explore Vision-Language models that understand night scenes and is efficient for on-device ML to enable deployment of women safety AI to mainstream smartphones.

To the best of our knowledge, some of the contributions in this paper are near first in the literature.

Key contributions are
1) The paper proposed & demonstrated a Deep Neural Network capable of translating low brightness images into sentences that describe the night scene. Experiments on the trained model shows this AI is able to caption any low light image that is randomly downloaded from the internet. While further research is required, this paper establishes the potential of using Deep Neural Network based Vision-Language modelling for on-device ML for night scene perception challenges. Thus, the paper establishes the potential for AI in safety of women & visual impaired users, especially at night.
2) A new concept of Interactive Image Captioning was explored & demonstrated. This concept allowed for usage model similar to Visual Question Answering (VQA). The user inputs both an image and a question in VQA. The proposed Interactive Captioning AI achieves similar user experience, where user inputs both an image and a question. This paper experimentally demonstrates this concept with integrating Human interaction into the Attention score layer of the Neural Network. Thus, when the user wishes to focus on a particular object of interest in the scene, she can prompt the Attention mechanism to consider her wish for caption generation. This interesting Human-in-the loop approach in the Attention-Decoder allows the user to influence the attention scoring weights.
3) To enable reproducibility of results, the paper contributes the source code for both the contributed ideas in open-source. The source code is made available at this URL , https://sites.google.com/view/lowlightimagecaption. A paper also makes a beautiful illustration of the neural network architecture, to support in-depth exploration by interested researchers.

The potential of Deep Learning to understand night scenes was demonstrated in this work. This paper opened door for new possibilities for applications to promote the safety of visually impaired women. Using Attention modelling, the paper developed a Encoder-Attention-Decoder based model that learns to interactively caption images of low brightness. Thus an Interactive Image Captioning AI for Explainable night scene understanding is demonstrated experimentally. To encourage further research by interested researchers in Vision-Language modelling for promoting safety of women, the source code is made available in open-source.